\documentclass[conference, a4paper]{IEEEtran}
\IEEEoverridecommandlockouts

\usepackage{epsfig}
\usepackage{balance}

\def\BibTeX{{\rm B\kern-.05em{\sc i\kern-.025em b}\kern-.08em
    T\kern-.1667em\lower.7ex\hbox{E}\kern-.125emX}}

\begin{document}

\title{Preliminary Forensics Analysis of DeepFake Images\\}

\author{\IEEEauthorblockN{Luca Guarnera}
\IEEEauthorblockA{\textit{Department of Mathematics}\\ \textit{and Computer Science} \\
\textit{University of Catania}\\
Catania, Italy \\
luca.guarnera@unict.it} 
\and
\IEEEauthorblockN{Oliver Giudice}
\IEEEauthorblockA{\textit{Department of Mathematics}\\ \textit{and Computer Science} \\
\textit{University of Catania}\\
Catania, Italy \\
giudice@dmi.unict.it} \\
\and
\IEEEauthorblockN{Cristina Nastasi}
\IEEEauthorblockA{\textit{Department of Mathematics}\\ \textit{and Computer Science} \\
\textit{University of Catania}\\
Catania, Italy \\
cristina.nastasi@unict.it}
\and
\IEEEauthorblockN{Sebastiano Battiato}
\IEEEauthorblockA{\textit{Department of Mathematics}\\ \textit{and Computer Science} \\
\textit{University of Catania}\\
Catania, Italy \\
battiato@dmi.unict.it}
}

\maketitle

\begin{abstract}
One of the most terrifying phenomenon nowadays is the Deepfake: the possibility to automatically replace a person's face in images and videos by exploiting algorithms based on deep learning. This paper will present a brief overview of technologies able to produce Deepfake images of faces. A forensics analysis of those images with standard methods will be presented: not surprisingly state of the art techniques are not completely able to detect the fakeness. To solve this, a preliminary idea on how to fight Deepfake images of faces will be presented by analysing anomalies in the frequency domain.
\end{abstract}


\section{Introduction}
\label{sect:introduction}

\begin{figure}[t]
    \centering
    \includegraphics[width=7.6cm]{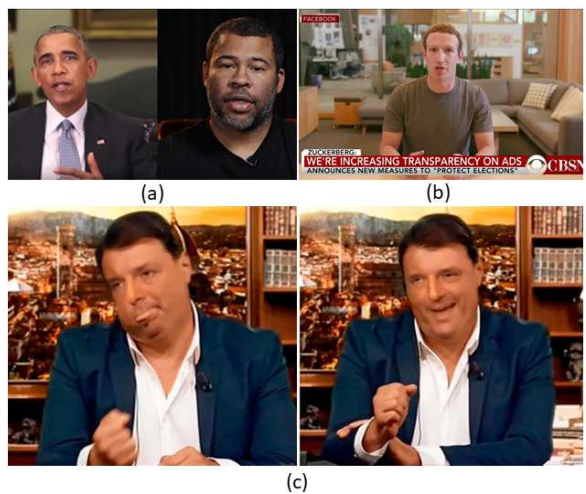}
    \caption{Several examples of DeepFake: (a) Obama, created by Buzzfeed in collaboration with Monkeypaw Studios; (b) Mark Zuckerberg, created by artists Bill Posters and Daniel Howe in partnership with advertising company Canny; (c) Matteo Renzi, created by ``Striscia la Notizia”.}
    \label{fig:1}
\end{figure}

Artificial intelligence technologies~\cite{plebe2019unbearable} are evolving so rapidly that unthinkable new applications and services have emerged: one of them is the DeepFake.
DeepFakes refers to all those multimedia contents synthetically altered or created by exploiting machine learning generative models. DeepFakes are image, audio or video contents that appear extremely realistic to humans specifically when they are used to generate and/or alter/swap image of faces. 
Various examples of DeepFake, involving celebrities, have already be seen on the internet: the insertion of Nicholas Cage\footnote{https://www.youtube.com/watch?v=-yQxsIWO2ic} in movies where he did not act like ``Fight Club" and ``The Matrix" or the impressive video in which Jim Carrey\footnote{https://www.youtube.com/watch?v=Dx59bskG8dc} plays Shining in place of Jack Nicholson. Other more worrying examples are the video of Obama (Figure~\ref{fig:1}(a)), created by Buzzfeed\footnote{
https://www.youtube.com/watch?v=cQ54GDm1eL0} in collaboration with Monkeypaw Studios, or the video in which Mark Zuckerberg\footnote{https://www.youtube.com/watch?v=NbedWhzx1rs} (Figure~\ref{fig:1}(b)) claims a series of statements about the platform's ability to steal its users' data. 

These Deepfakes have already been spread by mass media, also in Italy, where the satirical tv program ``Striscia La Notizia” \footnote{https://www.striscialanotizia.mediaset.it/video/}, broadcasted in September 2019 a video of the ex-premier Matteo Renzi talking about his colleagues in a ``not so respectful” way (Figure~\ref{fig:1} (c)). As we can imagine, DeepFakes may have serious repercussions on the veracity of the news spread by the mass media while representing  a new threat for politics, companies and personal privacy. 

Deepfakes are evolving quickly and are becoming dangerous, not just for the reputation of the victims but also for security. 
In this dangerous scenario, tools are needed to unmask the deepfakes detect them or, at least, to mitigate the potential harm and abuse that can be done be means of these multimedia contents.

Several big companies, from Facebook to Microsoft, have decided to take action against this phenomenon: 
Google has created a database of fake videos~\cite{rossler2019faceforensics++} to support researchers who are developing new methods to detect them while Facebook and Microsoft have launched the Deepfake Detection Challenge initiative\footnote{https://deepfakedetectionchallenge.ai/} which invites people from all over the world to create new tools to detect deepfakes and manipulated media.

The motivation of this paper is straightforward, after a brief overview of the state of the art in order to better understand what the technologies able to produce Deepfake are, a preliminary forensics analysis will be carried out. A first contribution to the field will be demonstrating that it is possible for the image forensics expert to find anomalies that could be related to how the Deepfake image is made. In fact standard image forensics tools are able to highlight some anomalies which the expert can analyze in deep to find specific anomalies in the frequency domain. In particular the anomalies, after Fourier transform, will be shown and will make clear that each kind of DeepFake creation technology has an easily detectable pattern. These evidence - as a preliminary result to the field - could lead to further and more sophisticated (even automatic) detection and analysis techniques.

The remaining part of the paper is organized as follow. Section~\ref{sect:SA} presents an overview of DeepFake creation technologies. Section~\ref{sect:Detection} investigates how Image Forensics can fights DeepFakes. The proposed method is described in Section~\ref{sect:Analysis}. Section~\ref{sect:conclusion} concludes the paper with the explanation of our future goal in this field.

\section{An overview of Generative technologies}
\label{sect:SA}
\subsection{Brief introduction to GAN}
Synthetic audiovisual media can be generated with a variety of techniques.
An overview on Media forensics with particular focus on Deepfakes has been recently proposed in~\cite{verdoliva2020media, tolosana2020deepfakes}.
Currently, the most popular of these techniques is the Generative Adversarial Network (GAN) due to its flexible applications and realistic outputs. Generative adversarial networks (GANs) were firstly introduced by Ian Goodfellow \cite{goodfellow2014generative} in 2014. They propose a new framework for estimating generative models via an adversarial process, in which we simultaneously train two models: a generative model $G$ that captures the data distribution, and a discriminative model $D$ able to estimate the probability that a sample came from the training data rather than $G$. The training procedure for $G$ is to maximize the probability of $D$ making a mistake. This framework corresponds to a min-max two-player game. 
In the case of Deepfakes, the $G$ can be thought as a team of counterfeiters trying to produce fake currency, while the $D$ stands to the police, trying to detect the malicious activity. $G$ and $D$ can be made by any kind of generative model, in the original version they are implemented through deep neural networks.


There are many types of models proposed in the literature \cite{karras2017progressive,radford2015unsupervised,zhang2017stackgan,zhu2017unpaired}. 
Architecture variant GANs are mainly proposed for the purpose of different applications e.g., image completion~\cite{iizuka2017globally}, image super resolution~\cite{ledig2017photo}, text-to-image generation~\cite{reed2016generative} and image to image transfer~\cite{zhu2017unpaired}. The original GAN paper~\cite{goodfellow2014generative} employed fully-connected neural networks for both generator and discriminator. 
\textit{Laplacian Pyramid of Adversarial Networks} is proposed for the production of higher resolution images from lower resolution input GAN~\cite{denton2015deep}. 
\textit{Deep Convolutional GAN} is the first work where a deconvolutional neural networks architecture~\cite{zeiler2014visualizing} is applied. 
\textit{Boundary Equilibrium GAN} uses an autoencoder architecture for the discriminator which was first proposed in~\cite{zhao2016energy}. 
\textit{Progressive GAN} involves progressive steps toward the expansion of the network architecture~\cite{karras2017progressive}. This architecture uses the idea of progressive neural networks first proposed in~\cite{rusu2016progressive}. 
\textit{BigGAN}~\cite{brock2018large} has also achieved state-of-the-art performance on the ImageNet datasets.

\subsection{Technologies for image creation}
Focusing on DeepFake images of face, in~\cite{lample2017fader} Lample et al. an ``encoder-decoder” architecture, (\textit{Fader Networks}), able to generate different realistic versions of an input image by varying the values of the attributes was introduces: given an input image $x$ with its attributes $y$, the encoder maps $x$ to a latent representation $z$, and the decoder is trained to reconstruct $x$ given $(z,y)$. At inference time, a test image is encoded in the latent space and the user chooses the values of the attributes $y$ that are sent to the decoder. A classifier learns how to predict the $y$ attributes given the latent representation $z$ during training. The encoder-decoder is trained so that the latent representation $z$ must contain enough sufficients information to allow input’s reconstruction while the latent representation must prevent the classifier from predicting the correct attribute values. The authors have trained and tested Fader Network considering the CelebA dataset~\cite{liu2015deep}, obtaining a model that can significantly change the perceived value of the attributes while preserving the natural appearance of the input images. Tests have also been carried out on the Oxford-102 dataset~\cite{nilsback2008automated} (containing about $9,000$ images of flowers classified in $102$ categories), by changing the colour of the flower while keeping the background unchanged. Excellent results have also been achieved in this test. 
The code is available at https://github.com/facebookresearch/FaderNetworks.

In~\cite{shen2017learning}, Shen et al. a novel method based on \textit{residual image learning} for face attribute manipulation is proposed. It can model the manipulation operation, as learning the residual image, defined as the difference between the original input image and the desired manipulated one. The proposed work focuses on the attribute-specific face area instead of the entire face which contains many redundant and irrelevant details. They develop a dual scheme able to learn two inverse attribute manipulations (one as the primal manipulation and the other as the dual manipulation) simultaneously. 
For each face attribute manipulation there are two image transformation networks called $G_0$ and $G_1$ and a discriminative network $D$. $G_0$ and $G_1$ that respectively simulate the primal and the dual manipulation. $D$ classifies the reference images and generated images into three categories. 

Several DeepFake based techniques that are present at the state of the art are limited regarding the management of more than two domains (for example, to change hair color, gender, age, and many others features in a face), since they should generate different models for each pair of image domains. A method capable of performing image-to-image translations on multiple domains using a single model has been proposed by Choi et al.~\cite{choi2018stargan} by means of a technology called \textit{StarGAN}, a generative adversarial network. The main purpose was to define a scalable image-to-image translation model across multiple domains using a single generator and a discriminator. The authors used two different types of face datasets: CelebA~\cite{liu2015deep} 
and RaFD dataset~\cite{langner2010presentation} 
Given a random label (for example hair color, facial expression, etc) as input, the network is able to perform an image-to-image translation operation considering the given label. 
The results have been compared with other existing methods~\cite{li2016deep, perarnau2016invertible, zhu2017unpaired} and show how StarGAN manages to generate images of superior visual quality.
The code is available at https://github.com/yunjey/stargan.

Wang et al.~\cite{wang2018face} propose Identity-Preserved Conditional Generative Adversarial Networks (\textit{IPCGAN}), a framework for facial aging. IPCGAN is composed of three parts: a CGANs module (takes an input image and a target age to generate a new face with that age), an identity-preserved module (guarantees the aged face has the same input identity) and an age classifier (to ensure that the output has the desired age). 
The authors considered faces with different ages divided into 5 groups: 11-20, 21-30, 31-40, 41-50 and 50+. Given an image of the face $x$, they use the $C_s$ information to indicate the age group to which $x$ belongs. The aging of a face aims to generate a synthesized face of the target age group $C_t$. The framework has been trained and tested using the Cross-Age Celebrity Dataset (CACD)~\cite{chen2014cross}. The performances of IPCGAN have been compared with acGAN~\cite{antipov2017face} and CAAE~\cite{zhang2017age} which achieve performance at vanguard for aging of the face. The qualitative and quantitative tests show that IPCGAN achieves the best results.  Finally, IPCGAN can also be used to perform multi-attribute transfer tasks. The code is available at https://github.com/dawei6875797/Face-Aging-with-Identity-Preserved-Conditional-Generative-Adversarial-Networks.

The Style Generative Adversarial Network, or \textit{StyleGAN}~\cite{karras2019style}, proposes large changes to the generator model, including the use of a mapping network to map points in latent space to an intermediate latent space, to control style at each point in the generator model, and the introduction to noise as a source of variation at each point in the generator model. The resulting model is capable not only of generating impressively photorealistic high-quality photos of faces, but also offers control over the style of the generated image at different levels of detail through varying the style vectors and noise. In December 2018, the visual computing company NVIDIA, released an open source code for photorealistic face generation software created thanks to the StyleGAN algorithm~\cite{karras2019style}. Then, Uber's computer engineer Phillip Wang created the website~https://thispersondoesnotexist.com/ and then on 11 February 2019 published it on the public group Facebook Artificial Intelligence and Deep Learning. StyleGAN algorithm is to be able to create realistic pseudo-portraits, difficult to judge as fakes.
StyleGAN has also difficulty with the definition of the teeth and it cannot identify the backgrounds. Furthermore, there are often fluorescent spots, similar to water drops, which can appear anywhere on the image. To correct those imperfections in StyleGAN, Karras et al. made some improvements to the generator (including re-designed normalization, multi-resolution, and regularization methods) proposing StyleGAN2~\cite{karras2019analyzing}.

\section{Fighting DeepFakes with Image Forensics}
\label{sect:Detection}
Image forgery and alteration is not a new problem introduced by DeepFakes. Counterfeiting an image with image editing tools like Photoshop is still very common as today and the image/multimedia forensics science as already dealt with the problem~\cite{battiato2016multimedia}. There are several techniques that try to understand if a multimedia content is fake by means of various strategies to detect anomalies in the hidden structure of the multimedia content itself, exploiting noise, compression parameters, etc. Some try to reconstruct the history of the image~\cite{giudice2017classification} to identify the source-acquiring device or software, others instead analyze anomalies in the compressed JPEG domain~\cite{battiato2009digital, giudice20191, galvan2013first} like Galvan et al.~\cite{galvan2013first} who proposed a method which is able to recover the coefficients of the first compression process in a double compressed JPEG image to verify if there are altered elements. Another example from Battiato et al.~\cite{battiato2009digital} exploits the statistical distribution of the DCT coefficients in order to detect irregularities due to the presence of a signal overlapped on the original image. More recently, Giudice et al.~\cite{giudice20191} proposed a new analysis that can be carried out in the DCT domain able to automatically classify doubly compressed JPEG images with extremely high precision, giving forensics experts a tool to find the first evidence of image alteration.
Not only the hidden structure of the image can be useful fakeness analysis. Even if the DeepFake images are extremely realistic, the visible contents could be analysed in order to find anomalies useful for detection. 

\begin{figure*}[t]
    \centering
    \includegraphics[width=17.2cm]{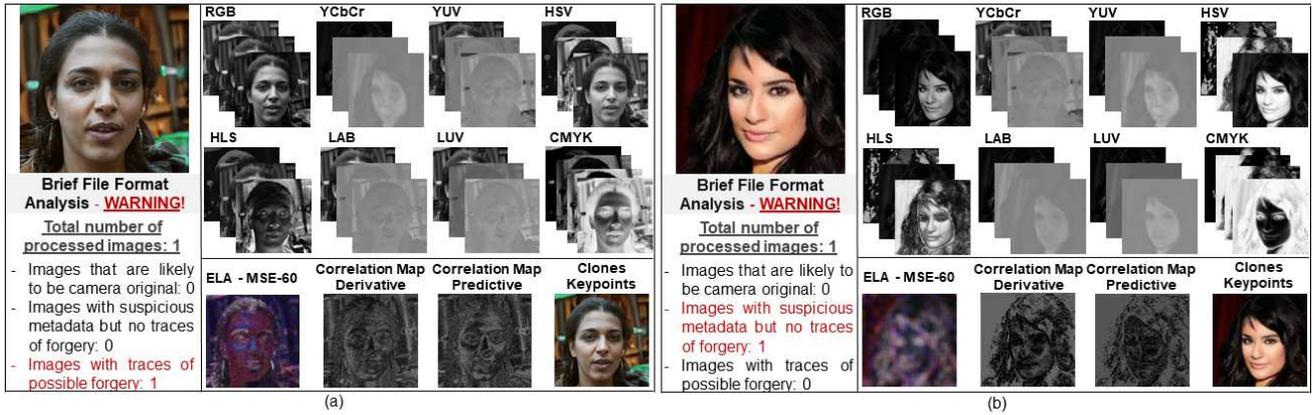}
    \caption{Example of Analysis carried out with Amped Authenticate software. (a) Image generated by STYLEGAN. (b) Image generated by STARGAN. In both examples we show only some of the elements analyzed with Amped Authenticate, such as the analyzed different color spaces to understand if anomalies are found; ELA (Error Level Analysis) for Identification of spliced areas of the image that have been compressed differently; Correlation map to analyze and identify the correlation between the pixels of the image; Clones Keypoints to find parts of the image that appear to be cloned.}
    \label{fig:Authenticate}
\end{figure*}

In this context, Marra et al.~\cite{marra2018detection} discussed the performance of various image-to-image translations detectors, both in ideal conditions and in the presence of compression, performed at the time of uploading to social networks. The study, conducted on a dataset of $36.302$ images, shows that it is possible to obtain detection accuracy up to $95\%$ both with conventional detectors and with deep learning based detectors, but only the latter continue to provide high accuracy, up to $89\%$, on compressed data. 

Hsu et al.~\cite{hsu2018learning} also proposed an interesting method for detecting Deepfakes, called Deep Forgery Discriminator (DeepFD). Thanks to the implementation of a new discriminator that uses ``contrastive loss” it is possible to find the typical characteristics of the synthesized images generated by different GANs and therefore use a classifier to detect such fake images. For the training phase they used the CelebA dataset~\cite{liu2015deep}, considering 5 state-of-the-art GANs to generate the pool of false images~\cite{arjovsky2017wasserstein, gulrajani2017improved, karras2017progressive, mao2017least, radford2015unsupervised}. 
Using the DeepFD they detected $94.7\%$ of fake images generated by numerous state-of-the-art GANs, exceeding the other basic approaches present in the state of the art in terms of precision and recall rates. The code is available at https://github.com/jesse1029/Fake-Face-Images-Detection-Tensorflow.

As described by Guarnera et al.~\cite{guarnera2020deepfake}, the current GAN architectures that create Deepfake images, through convolution layers, leaves a fingerprint that characterizes that specific neural architecture. In order to capture this forensic trace, the authors used the Expectation-Maximization Algorithm~\cite{moon1996expectation} obtaining features able to distinguish real images from Deepfake ones.

Wang et al.~\cite{wang2019cnngenerated} investigated the possibility to create an universal detector able to identify the real images generated by a CNN regardless of the architecture or dataset used. They trained a classifier on a single CNN generator (ProGAN~\cite{karras2017progressive}). The conducted experiments demonstrated that this classifier detects synthesized images generated by different architectures and is also robust to JPEG compression, spatial blurring and scaling.

There are some specific techniques also for video which try to define if they are fakes. A video based DeepFake Detection method is described in the work of Güera et al.~\cite{guera2018deepfake}. The authors used a CNN to extract frame-level features. In particular they use a Recurrent neural network (RNN) to train a classifier able to recognise if the video was manipulated or not by evaluating temporal inconsistencies introduced between the frames where face is modiﬁed. They evaluated the method with $600$ videos: $300$ fakes found on the web, $300$ of real scenes taken from the HOHA dataset~\cite{laptev2008learning}. From experiments carried out by the authors, they get $97\%$ accurate for fake detection.

Finally, Rossler et al.~\cite{rossler2019faceforensics++} proposed an automated benchmark for fake detection, based mainly on four manipulation methods: two computer graphics-based methods (Face2Face~\cite{thies2016face2face}, FaceSwap\footnote{https://github.com/MarekKowalski/FaceSwap/}) and 2 learning-based approaches (Deepfakes\footnote{https://github.com/deepfakes/faceswap/}, NeuralTextures~\cite{thies2019deferred}). 
The authors addressed the problem of fake detection as a binary classification for each frame of manipulated videos, considering different techniques present in the state of the art ~\cite{afchar2018mesonet, bayar2016deep, chollet2017xception, cozzolino2017recasting, fridrich2012rich, rahmouni2017distinguishing}. 

A face tracking method is initially applied to the input image~\cite{thies2016face2face} 
(so as to work only on that particular region) to switch to the classification methods. The experiments conducted show that XceptionNet achieves the best results. They also tested XceptionNet with images without face tracking information. In this case, however, the XceptionNet classifier has significantly lower accuracy. 
The code is available at https://github.com/ondyari/FaceForensics.

\section{DeepFake Forensics Analysis}
\label{sect:Analysis}
In the previous paragraphs the most accurate techniques used nowadays to create DeepFake were described, and therefore different detection techniques were discussed. Most of the existing detection techniques are based on neural networks and it is very complicated to explain their results in a courtroom, giving their black-box features. Being able to understand the type of architecture used, the anomalous features and deterministically explaining why an image is a Deepfake is a process that is still almost completely absent in the literature.

\begin{figure}[t]
    \centering
    \includegraphics[width=7.6cm]{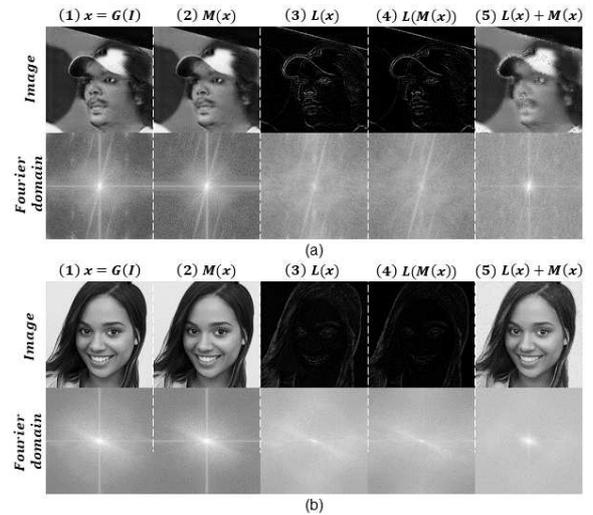}
    \caption{Examples of analyzed images generated by (a) STARGAN and (b) STYLEGAN. Each image $I$ of both datasets was converted to grayscale (1) and applied progressively: the Median filter (2), the Laplacian filter (3), the Laplacian filter (4) applied to the result of (2), the sum of the results between the Median and Laplacian filters (5). For each operation performed, we show the Fourier transform.}
    \label{fig:LaplacianMedia}
\end{figure}

\setlength{\belowcaptionskip}{-20pt}

In general, in the Multimedia Forensics~\cite{battiato2016multimedia} best practices are used to determine if an image is a fake, where, one tries to understand if it comes from a particular source, analyzing, information present in the metadata (how much present and not altered), the PNRU present in the images (fingerprint of the source), analysis of the coefficients obtained by JPEG compression, analysis in the Fourier domain and much more.
If a specific information is not present, the suspicion arises that the data in question is not real. Nowadays, the current neural networks that generate DeepFake do not perform the same operations that are performed by any acquisition imaging source. This leads to obtain images with anomalies at the pixel level, obtaining a pattern that is not present if we consider an image generated by any device or software (camera, scanner, social network, etc.).

A forensics analysis was carried out on sample Deepfake Images by means of one of the most famous image forensics software ``Amped Authenticate"\footnote{https://ampedsoftware.com/it/authenticate/}: it was employed to check if it is possible to identify whether an image, generated by a GAN, has anomalies, considering in particular the output of two types of technologies: StarGAN~\cite{choi2018stargan} and StyleGAN~\cite{karras2019style}, briefly described Section~\ref{sect:SA}. In particular we analyzed JPEG structure of the image, we tried to infer Camera Identification (PNRU Identification), and then analysed those images in different color spaces (RGB, YCbCr, YUV, HSV, HLS, XYZ, LAB, LUV, CMYK); domains (ELA, DCT Map, JPEG Dimples Map, Blocking Artifacts, JPEG Ghosts Map, Fusion Map, Correlation Map, PRNU Map, PRNU Tampering, LGA) and by means of many forgery detection techniques (Clones Blocks, Clones Keypoints (Orb and Brisk). Figure~\ref{fig:Authenticate} shows an overview of the results obtained and it is possible to see that in some cases the images may show specific anomalies while in other cases simple warnings are shown by the tool. However, this is not enough to define with certainty if the images are Deepfake, the only thing that could be inferred is that they are probably not-authentic and integrity is broken. 

A deep analysis was then carried out in the frequency domain. Indeed, useful information can be obtained by working on Deepfake “candidate" images after the Fourier transform. In general, a simple operation of Forgery or a Deepfake contains “abnormal” frequencies not present in real images. The application of convolutive filters with the respective Fourier spectra highlights the presence of a somewhat suspicious pattern, not present in real data. We performed several tests, using the Laplacian and median filters and the combination of them in order to enhance them (Figure~\ref{fig:LaplacianMedia}). In the Fourier domain, as shown in Figure~\ref{fig:LaplacianMedia}, it is possible to notice anomalous frequencies that substantially represent a pattern of that particular network used to generate fakes. This information is useful for identifying the type of neural network used and the areas in which that pattern is present. In Figure~\ref{fig:LaplacianMedia} the anomalies are clearly visible and are different for the two samples of each different techniques. Probably these patterns represents the way that deep neural network, and their convolutive layers, create the image, so they could be related to the hyper-parameters such as kernel masks employed in each generative technique. Obviously further investigation is needed but the conjunction of standard image forensics techniques that arises warning of ``fakeness" and the detection of known anomalies (different for each technology) in the Fourier domain can achieve good results in terms of Deepfake Detection performance.

Forensics vs. anti-forensics is always an open game, and while we are dealing with detection methods, there are already a attempts seeking to hide those anomalies that were described above by introducing new camuoflage information into the fake images. This method was already proposed by Cozzolino et al.~\cite{cozzolino2019spoc}, in his work called ``SpoC: Spoofing Camera Fingerprints” in which they proposed a GAN-based approach capable of injecting traces of the camera fingerprint into images (thus probably reducing traces of the synthesis process), thus tricking avant-garde detectors.


\section{Conclusion}
\label{sect:conclusion}
In this paper we presented several techniques of creation and detection of the so-called Deepfakes and the related social and legal problems. It turns out to be very important to be able to create new methods that can counter this phenomenon. This could be done by analyzing details and traces of underlying generation process of the image (e.g. in the Fourier domain).

\label{sect:bib}
\balance
\bibliographystyle{IEEEtran}
\bibliography{main}

\end{document}